\documentclass[10pt,twocolumn,letterpaper]{article}

\usepackage{cvm}
\usepackage{times}
\usepackage{epsfig}
\usepackage{graphicx}
\usepackage{amsmath}
\usepackage{amssymb}
\usepackage{booktabs}
\usepackage{multirow}
\usepackage[ruled,vlined]{algorithm2e}  
\usepackage{algorithmic}     
\usepackage{float}

\usepackage[pagebackref=true,breaklinks=true,colorlinks,bookmarks=false]{hyperref}

\newcommand{\keywords}[1]{{\bf \emph{Keywords: #1}}}

\cvmfinalcopy

\def\cvmPaperID{****} 

\ifcvmfinal\fi
\begin{document}

\title{Residual Cross-Modal Fusion Networks for Audio-Visual Navigation}

\author{
Yi Wang \qquad Yinfeng Yu\thanks{Yinfeng Yu is the corresponding author (Email: yuyinfeng@xju.edu.cn)}\\
School of Computer Science and Technology, Xinjiang University, Urumqi, China\\
Joint International Research Laboratory of Silk Road Multilingual Cognitive\\ 
%
\and
%
Bin Ren\\
School of Mechatronic Engineering and Automation\\
Shanghai University, Shanghai, China
}

\maketitle

\begin{abstract}
    Audio-visual embodied navigation aims to enable an agent to autonomously localize and reach a sound source in unseen 3D environments by leveraging auditory cues. The key challenge of this task lies in effectively modeling the interaction between heterogeneous features during multimodal fusion, so as to avoid single-modality dominance or information degradation, particularly in cross-domain scenarios. To address this, we propose a Cross-Modal Residual Fusion Network, which introduces bidirectional residual interactions between audio and visual streams to achieve complementary modeling and fine-grained alignment, while maintaining the independence of their representations. Unlike conventional methods that rely on simple concatenation or attention gating, CRFN explicitly models cross-modal interactions via residual connections and incorporates stabilization techniques to improve convergence and robustness. Experiments on the Replica and Matterport3D datasets demonstrate that CRFN significantly outperforms state-of-the-art fusion baselines and achieves stronger cross-domain generalization. Notably, our experiments also reveal that agents exhibit differentiated modality dependence across different datasets. The discovery of this phenomenon provides a new perspective for understanding the cross-modal collaboration mechanism of embodied agents.
\end{abstract}


\section{Introduction}

With the rapid development of embodied intelligence~\cite{embodied1,embodied2,embodied3,embodied4,yu2025dope}, enabling agents to accomplish complex tasks in real or simulated environments has become a critical research direction in the field of artificial intelligence~\cite{visualN1,visualN2}. Among related tasks, audio-visual navigation (AVN) is a key task for embodied intelligence, which requires agents to autonomously locate and navigate to target sound sources in unknown environments by leveraging both visual information and auditory cues. Such tasks hold broad value in application scenarios, including service robotics, human-computer interaction, and security monitoring, thus attracting increasing attention.

\begin{figure}
    \centering
    \includegraphics[width=1\linewidth]{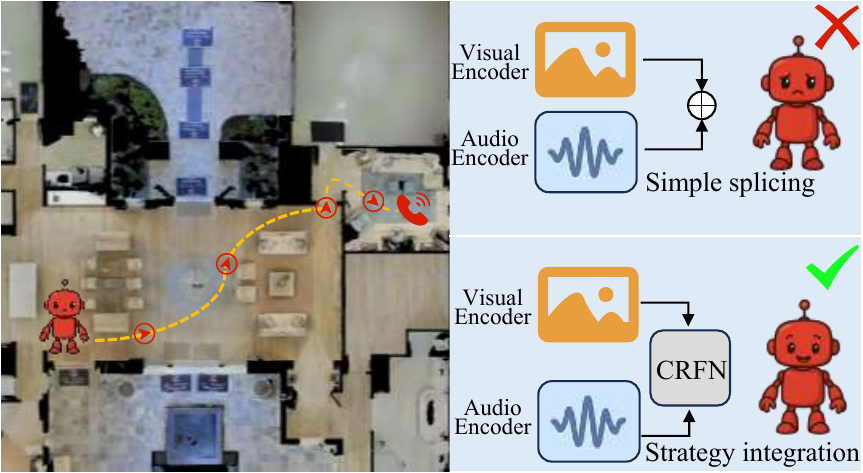}
    \caption{Humans can effortlessly integrate visual and auditory information, yet agents often suffer from performance degradation due to modal imbalance or domain discrepancy.}
    \label{T-fig}
\end{figure}

In a bustling, noisy shopping mall, you faintly hear someone calling your name. Your ears first pick up the sound's general direction; you then turn your head and scan the crowd with your gaze. Only when you catch sight of a familiar figure do you confirm the caller’s position. Amid this chaotic setting, hearing narrows down the search range for you, while vision finalizes the identification and locking in of the target. The two work in seamless harmony, requiring hardly any deliberate effort on your part.

\begin{figure*}
    \centering
    \includegraphics[width=1\linewidth]{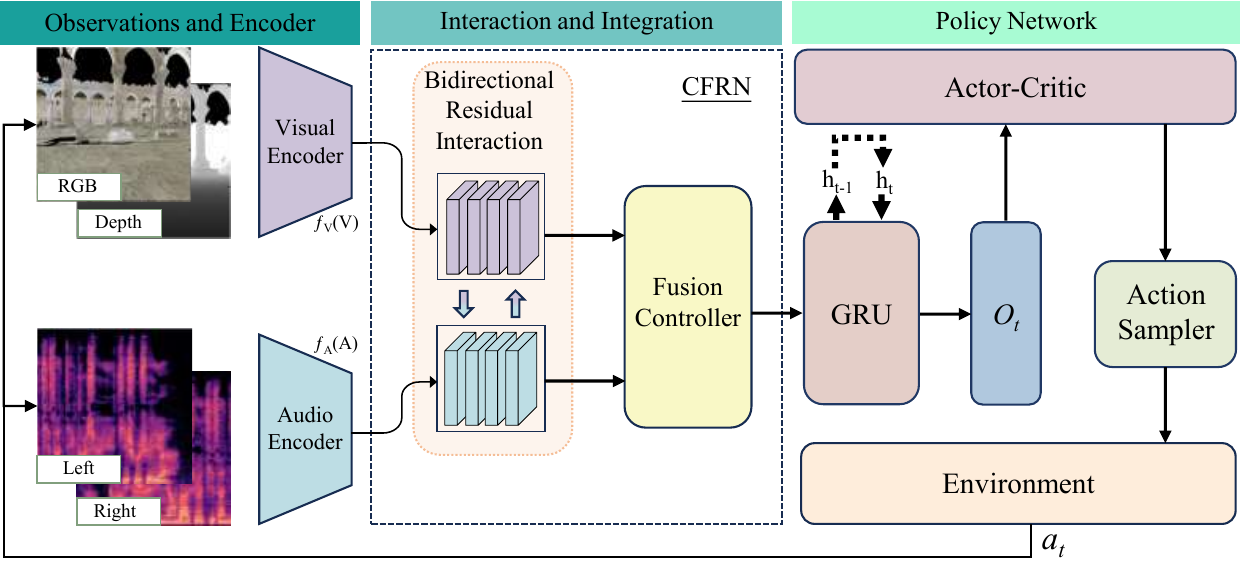}
    \caption{\textbf{Illustration of the basic architecture}. The agent's navigation process operates in three stages: (1) Observation and Encoder, where the agent processes visual (RGB or Depth) and auditory (spectrogram) inputs via respective encoders to extract features; (2) Interaction and Integration, where our proposed CRFN module performs bidirectional residual updates to refine features and adaptively balances them using a Fusion Controller; and (3) Policy Network, where the fused representation is fed into a GRU-based Actor-Critic model to capture temporal dependencies and predict the final navigation action $a_t$.}
    \label{fig1}
\end{figure*}

This kind of sensory synergy, which seems effortless to humans, poses a significant challenge for embodied agents: while these agents can acquire both visual and auditory information, the two modalities often operate in isolation, lacking genuine complementarity and interaction---this leads to a substantial decline in navigation performance in complex environments. On the other hand, an interesting phenomenon was also observed in the experiments of this paper: the fusion effect exhibits significant differences across different environments. In synthetic high-fidelity scenes (e.g., Replica~\cite{replica}), the visual modality often becomes dominant due to its clearer texture and geometric information, gradually weakening the audio modality. In contrast, in complex real-world scenes (e.g., Matterport3D~\cite{matterport3d}), a single modality can hardly complete the navigation task independently, and cross-modal complementarity instead becomes the key to performance improvement. This dynamic shift between modality dominance and cross-modal complementarity reveals the instability of existing methods in cross-environment and cross-domain generalization, and further underscores the necessity of designing a robust fusion mechanism.

To address these challenges, this paper introduces a novel Cross-Modal Residual Fusion Network (CRFN). The proposed method is built upon two core design principles: bidirectional residual interaction and a fusion control mechanism. Specifically, reciprocal residual pathways are established between audio and visual features to enable mutual refinement and complementarity while preserving the independence of each modality. In addition, a lightweight fusion controller is incorporated to adaptively regulate modality contributions during interaction, with an output normalization constraint applied to suppress single-modality dominance and stabilize training. This modular design not only enhances the robustness and generalization ability of the fusion process but also provides stronger interpretability: the evolution of residual coefficients offers an intuitive view into the dynamic collaboration between modalities.

In summary, our main contributions are as follows:

\begin{itemize}
\item \textit{A novel cross-modal fusion framework.} We propose the Cross-Modal Residual Fusion Network (CRFN), which introduces bidirectional residual interactions and a lightweight fusion controller to explicitly model reciprocal refinement between audio and visual modalities while preserving their independence.
\item \textit{Improved robustness and generalization.} By combining small-value initialization of residual scaling factors with output normalization, CRFN suppresses modality imbalance, stabilizes training, and consistently outperforms concatenation and attention-based baselines on both Replica~\cite{replica} and Matterport3D~\cite{matterport3d} benchmarks.
\item \textit{New empirical insights into modality dependence.} Through the analysis of residual coefficients, we uncover a novel phenomenon: in synthetic high-fidelity environments, the navigation policy tends to be dominated by the visual modality, whereas in real-world complex environments, cross-modal complementarity becomes essential. This provides a new perspective for understanding modality collaboration in embodied audio-visual navigation.

\end{itemize}

\begin{figure}
    \centering
    \includegraphics[width=\linewidth]{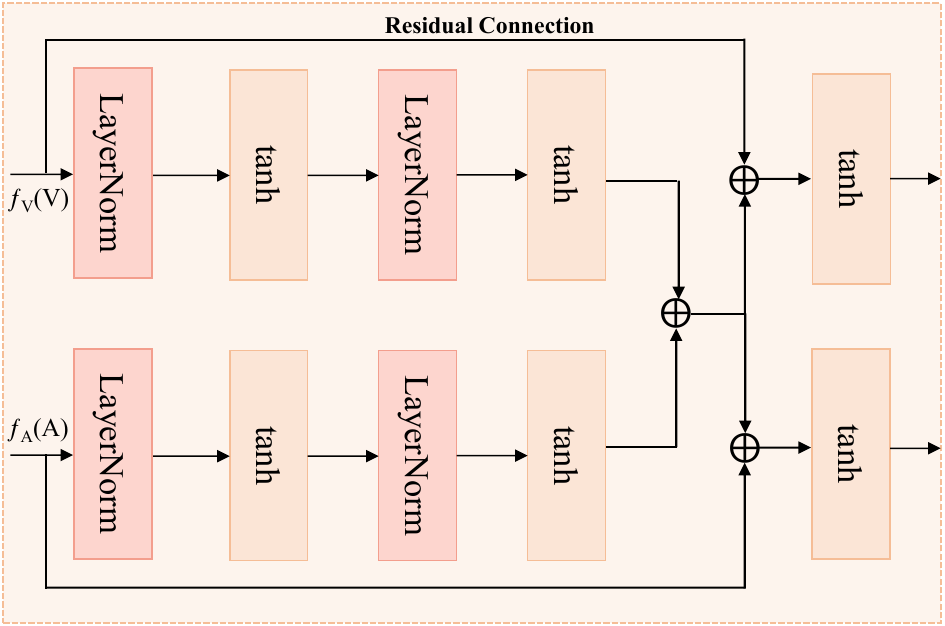}
    \caption{\textbf{Architecture of the Cross-Modal Feature Fusion Module.} This module enables visual and audio features to refine and complement each other through bidirectional residual paths. Features of each modality are updated and mutually influenced in the bidirectional interaction, ensuring a balanced exchange of information.}
    \label{fig:placeholder}
\end{figure}

\section{Related Work}
Embodied intelligence aims to enable agents to learn and accomplish complex tasks through interaction with the environment. Navigation is one of its core capabilities, and early research has primarily focused on visual navigation, such as PointGoal navigation~\cite{PointGN} and ObjectGoal navigation~\cite{ObjectGN}. In these tasks, goals are typically specified by visual or textual information, such as coordinates or object categories. While the aforementioned works have achieved significant progress in visual navigation, they limit the agent's perception to a single visual modality, which is far from the way humans and animals perceive the real world using multi-sensory synergy. In real physical environments, auditory perception provides a crucial information channel that complements vision. Sound signals possess the property of omni-directionality, allowing them to penetrate visual occlusions and provide key directional cues for targets outside the agent's field of view. To endow embodied agents with such more powerful and biologically intuitive perceptual and action capabilities, AVN has been proposed.

The formal definition of the AVN task and the establishment of large-scale benchmarks mark the starting point for the development of this field. Chen et al. have made pioneering contributions in this regard: based on the Habitat~\cite{habitat,habitat2.0} platform, they constructed SoundSpaces~\cite{soundspaces} on the Replica~\cite{replica} and Matterport3D~\cite{matterport3d} datasets, which is the first large-scale and high-fidelity audio-visual simulation environment and provides a unified experimental platform and testing benchmark for subsequent research. This work has established the core problems and evaluation criteria for the AVN field; however, its fusion method is relatively elementary, making it difficult to address modal imbalance in complex environments. In parallel, Gan et al. explored a decomposed modular framework, attempting to decouple the perception, localization, and planning components to enhance interpretability and scalability~\cite{gan2020look}. These two types of methods represent the two early mainstream approaches-end-to-end learning and modular planning-respectively, and together they have established the research baseline and development direction for the AVN field.

Early AVN frameworks exhibited low efficiency when handling long-range navigation tasks~\cite{yuttt}. To address this, AV-WaN~\cite{Waypoint} introduced an ``acoustic memory map" for the first time, which structurally records sound information perceived by the agent during movement. This work innovatively proposed a hierarchical reinforcement learning framework: the high-level policy dynamically predicts mid-to-long-term ``waypoints" based on multi-modal perception, while the low-level planner executes specific paths to reach these waypoints. Subsequently, SAVi~\cite{SAVi} further pointed out the limitations of existing tasks-namely, the assumption that sound sources emit continuously and lack semantic relevance. Accordingly, they proposed a more challenging semantic AVN task, where sounds are sporadic and transient, and semantically consistent with the category of objects producing the sounds (e.g., the flushing sound of a toilet). 

As fundamental issues have been addressed, researchers in the AVN field have shifted their focus toward more realistic and complex real-world scenarios. SAAVN~\cite{SAAVN,yu2023measuring,kim2016multimodal} took the lead in investigating the navigation robustness in acoustically complex environments: it designed an adversarial environment that incorporates a ``sound attacker," which actively moves and adjusts both volume and sound category to interfere with the navigating agent. Later, a benchmark for dynamic AVN was proposed, marking the first time an agent was required to track a moving sound source~\cite{catchme}. Almost simultaneously, FSAAVN~\cite{FSAVVN} also focused on the moving sound source tracking task and pointed out that the simple feature concatenation-based fusion methods used in previous works might overlook contextual information, and this was also the first time researchers paid attention to the issue of audio-visual cross-modal fusion since the inception of the AVN task. Subsequently, ORAN~\cite{ORAN} leveraged Cross-Task Curriculum Policy Distillation to transfer the knowledge of a pre-trained ``expert" policy (trained on the point-goal navigation task) to the AVN task. CAVEN~\cite{CAVEN} explored the role of human-computer interaction in AVN, introducing human interaction, large language models, and a new benchmark. RILA~\cite{RILA} made the first attempt to solve the AVN task in a zero-shot manner, completely eliminating the need for reinforcement learning training in the environment. Shi et al. contributed BeDAVIN, a large-scale audio dataset, as well as ENMUS³-an, an architecture specifically designed for multi-sound-source scenarios~\cite{tjavn}. 

Regrettably, although AVN is inherently a cross-modal task~\cite{yu2021weavenet}, the focus of existing research has largely been on innovations in navigation strategies and memory mechanisms. Its core audio-visual fusion module is to a large extent treated as a standardized component, and its inherent synergy potential has not been fully exploited.

\section{Proposed Method}

\subsection{Framework Overview}

As shown in the figure, the agent receives multimodal observations from the 3D environment, including visual images and audio signals. The visual modality consists of either depth maps or RGB images (one of which is used in the task setting), while the audio modality is represented as binaural spectrograms. First, visual and audio data are processed by dedicated encoders to extract modality-specific features, which are then fed into the proposed CRFN network. The primary role of CRFN is to establish bidirectional residual interactions and maintain a dynamic balance between modalities within a shared feature space, thereby producing robust fused representations. The fused features are subsequently passed into a recurrent unit (GRU) to capture historical information and temporal dependencies~\cite{srivastava2012multimodal}. Finally, an Actor-Critic policy head leverages the temporal representation to predict action distributions and state values, driving the agent to accomplish goal-directed navigation tasks.

To achieve robust and stable cross-modal fusion, the CRFN proposed in this paper consists of two main modules: a Modal Interaction Module and a Fusion Control Module. The Modal Interaction Module explicitly models the mutual interaction between visual and audio features through bidirectional residual paths, ensuring fine-grained modal complementarity and representation alignment. The Fusion Control Module, on the other hand, adaptively adjusts the contribution ratio of each modality via learnable residual scaling factors and normalization constraints, effectively suppressing the dominance of a single modality and stabilizing the training process. The former is responsible for establishing efficient information exchange channels, while the latter acts as a ``traffic controller" to dynamically regulate the intensity of such exchange, thereby achieving robust navigation performance in variable scenarios. CRFN adopts a concise and lightweight architecture, avoiding redundant network layers and complex structures to ensure efficient computation and training. With this design, the model can efficiently achieve cross-modal information fusion and deliver excellent performance in audio-visual navigation tasks. Sections~\ref{AA} and~\ref{BB} below will delve into the design details and implementation methods of these two modules, respectively.

\subsection{Bidirectional Residual Interaction}\label{AA}

This module aims to construct a symmetric information pathway, enabling the feature representations of the two modalities to complement and refine each other. Specifically, at each time step $t$, the visual features $v_t(f_v(V)$ and audio features $a_t(f_A(A))$ obtained through encoders first pass through their respective non-linear transformation networks $U_v(\cdot)$ and $U_a(\cdot)$ to be mapped to a shared representation space that facilitates interaction. Subsequently, we average the transformed features to generate an interaction vector $h_{interact}$ that integrates information from both modalities:\begin{equation}h_{interact} = \frac{1}{2} (U_v(v_t)+U_a(a_t)),\end{equation}This interaction vector $h_{interact}$ captures the core associations between audio and visual modalities and serves as a shared residual signal, which will be used to simultaneously update the features of both original modalities. This symmetric residual design ensures bidirectional flow of information between the two modalities.

\begin{algorithm}[t]
\caption{Fusion Controller in Cross-Modal Residual Fusion Network (CRFN)}
\label{alg:fusion_controller}
\KwIn{Visual feature $\mathbf{v}_t$, Audio feature $\mathbf{a}_t$, Interaction vector $\mathbf{h}_{interact}$}
\KwOut{Updated features $\hat{\mathbf{v}}_t$, $\hat{\mathbf{a}}_t$}
\textbf{Learnable parameters:} $\beta_v$, $\beta_a$ \\
\BlankLine
\textbf{Step 1: Normalize modal features}\\
$\mathbf{v}_{norm} \leftarrow \mathrm{LayerNorm}(\mathbf{v}_t)$ \\
$\mathbf{a}_{norm} \leftarrow \mathrm{LayerNorm}(\mathbf{a}_t)$ \\
\BlankLine
\textbf{Step 2: Compute residual updates}\\
$\mathbf{v}_{res} \leftarrow \mathbf{v}_{norm} + \beta_v \cdot \mathbf{h}_{interact}$ \\
$\mathbf{a}_{res} \leftarrow \mathbf{a}_{norm} + \beta_a \cdot \mathbf{h}_{interact}$ \\
\BlankLine
\textbf{Step 3: Apply activation to obtain final features}\\
$\hat{\mathbf{v}}_t \leftarrow \tanh(\mathbf{v}_{res})$ \\
$\hat{\mathbf{a}}_t \leftarrow \tanh(\mathbf{a}_{res})$ \\
\BlankLine
\Return{$\hat{\mathbf{v}}_t$, $\hat{\mathbf{a}}_t$}
\end{algorithm}

\begin{table*}[t]
\centering
\caption{Performance comparison with other methods under the Depth setting. SPL, SR, SNA are percentages.}
\label{tab:depth}

\renewcommand{\arraystretch}{1.3} 
\setlength{\tabcolsep}{6pt}       

\begin{tabular*}{\textwidth}{@{\extracolsep{\fill}} l|ccc|ccc|ccc|ccc @{}}
\toprule
& \multicolumn{6}{c|}{\textbf{Replica}} 
& \multicolumn{6}{c}{\textbf{Matterport3D}} \\
\cmidrule(lr){2-7} \cmidrule(lr){8-13}
\textbf{Method} &
\multicolumn{3}{c|}{\textbf{Heard}} 
& \multicolumn{3}{c|}{\textbf{Unheard sound}} 
& \multicolumn{3}{c|}{\textbf{Heard}} 
& \multicolumn{3}{c}{\textbf{Unheard sound}} \\
\cmidrule(lr){2-4} \cmidrule(lr){5-7} \cmidrule(lr){8-10} \cmidrule(lr){11-13}
& \textbf{SPL}$\uparrow$ & \textbf{SR}$\uparrow$ & \textbf{SNA}$\uparrow$
& \textbf{SPL}$\uparrow$ & \textbf{SR}$\uparrow$ & \textbf{SNA}$\uparrow$
& \textbf{SPL}$\uparrow$ & \textbf{SR}$\uparrow$ & \textbf{SNA}$\uparrow$
& \textbf{SPL}$\uparrow$ & \textbf{SR}$\uparrow$ & \textbf{SNA}$\uparrow$ \\
\midrule
Random~\cite{Waypoint}         & 4.9  & 18.5 & 1.8  & 4.9  & 18.5 & 1.8  & 2.1  & 9.1  & 0.8  & 2.1  & 9.1  & 0.8 \\
Direction Follower~\cite{Waypoint}   & 54.7 & 72.0 & 41.1 & 11.1 & 17.2 & 8.4  & 32.3 & 41.2 & 23.8 & 13.9 & 18.0 & 10.7 \\
Frontier W~\cite{Waypoint}   & 44.0 & 63.9 & 35.2 & 6.5  & 14.8 & 5.1  & 30.6 & 42.8 & 22.2 & 10.9 & 16.4 & 8.1 \\
Supervised W~\cite{Waypoint} & 59.1 & 88.1 & 48.5 & 14.1 & 43.1 & 10.1 & 21.0 & 36.2 & 16.2 & 4.1  & 8.8  & 2.9 \\
Gan et al.~\cite{gan2020look}           & 57.6 & 83.1 & 47.9 & 7.5  & 15.7 & 5.7  & 22.8 & 37.9 & 17.1 & 5.0  & 10.2 & 3.6 \\
SoundSpaces~\cite{soundspaces}          & 74.4 & 91.4 & \textbf{48.1} & 34.7 & 50.9 & 16.7 & 54.3 & 67.7 & 31.3 & 21.9 & 33.5 & 10.4 \\
\textbf{CRFN (Ours)} & \textbf{76.7} & \textbf{93.1} & 47.3 & \textbf{41.6} & \textbf{55.7} & \textbf{22.5} & \textbf{57.3} & \textbf{70.3} & \textbf{33.2} & \textbf{27.7} & \textbf{40.1} & \textbf{13.5} \\
\bottomrule
\end{tabular*}%
\end{table*}

\subsection{Fusion Controller}\label{BB}

The core of the fusion control module lies in adaptively adjusting modal contributions and ensuring the stability of the training process. This function is not implemented via a complex dynamic network, but rather accomplished by two independent, learnable scalar parameters—residual scaling factors $\beta _v$ and $\beta _a$. These two parameters control the update intensity of the interaction vector $h_{interact}$ on visual and audio features, respectively.

We substitute these two factors into the residual update path to obtain the updated features $\beta _v$ and $\beta _a$ :

\begin{equation}
    \hat{\mathbf{v}}_t = \text{act}(\text{LN}(\mathbf{v}_t) + \beta_v \cdot \mathbf{h}_{\text{interact}}),
\end{equation}

\begin{equation}
    \hat{\mathbf{a}}_t = \text{act}(\text{LN}(\mathbf{a}_t) + \beta_a \cdot \mathbf{h}_{\text{interact}}),
\end{equation}
Among them, $LN(\cdot)$ refers to layer normalization applied to the residual path, which is used to stabilize the input scale of each modality; $act(\cdot)$ denotes the Tanh activation function. At the start of training, $\beta _v$ and $\beta _a$ are initialized to a small value. This initialization encourages the model to start learning from a ``weakly coupled" state, which is more conducive for the network to independently discover the appropriate modal fusion intensity based on task requirements.

The overall computational procedure is summarized in Algorithm~\ref{alg:fusion_controller}, providing a concise representation of how the residual scaling factors $\beta_v$ and $\beta_a$ operate within the CRFN architecture.

\section{Experiments}

\subsection{Datasets and Implementation Detail}

Our experiments were conducted in a comprehensive simulated environment, which is built on the Habitat simulator~\cite{habitat} and SoundSpaces~\cite{soundspaces} acoustic platform, and leverages two complementary public 3D datasets: Matterport3D~\cite{matterport3d} and Replica~\cite{replica}. Among them, Matterport3D~\cite{matterport3d} is a large-scale, diverse dataset of real-world scans, containing 85 complex indoor environments. It serves as an ideal platform for evaluating the generalization ability of agents, but its models contain geometric noise inherent to real-world scanning. In contrast, Replica~\cite{replica} is a small yet high-quality dataset, comprising 18 high-fidelity synthetic indoor scenes. Renowned for its clean and realistic 3D meshes, Replica~\cite{replica} is highly suitable for validating the effectiveness of models in high-fidelity environments. Using the geometric and material information from these two datasets, the SoundSpaces~\cite{soundspaces} platform generates physically realistic sound sources by convolving 102 non-repetitive natural sounds with binaural Room Impulse Responses (RIRs) corresponding to specific directions.

Implementation Details. The model parameter configuration is as follows: the sampling rate of Room Impulse Response (RIR) is 44,100~Hz for the Replica~\cite{replica} dataset and 16,000~Hz for the Matterport3D~\cite{matterport3d} dataset. In all experiments, the resolution of both visual and auditory observations is $128 \times 128$. We adopt the Proximal Policy Optimization (PPO) algorithm for training, and its hyperparameters remain consistent across the two datasets: 4 training epochs, a clip parameter of 0.1, a value loss coefficient of 0.5, and a discount factor ($\gamma$) set to 0.99. For experiments on the Replica~\cite{replica} dataset, the learning rate is set to $2.5 \times 10^{-4}$, while for Matterport3D~\cite{matterport3d}, the learning rate is $2.0 \times 10^{-4}$. During training, the number of steps per episode is limited to 500, and the model undergoes 40,000 update iterations on the Replica dataset and 60,000 update iterations on the Matterport3D~\cite{matterport3d} dataset.

In our experiments, there are two distinct sound source conditions: (1) Heard: the target sound source is a telephone ringtone, which is used consistently across the training, validation, and test sets; (2) Unheard: the 102 sound sources are divided into three non-overlapping groups, where 78 sound sources are used for training scenarios, 11 for validation scenarios, and the remaining 18 for test scenarios. Every scenario in the test process is new to the model-none of them were encountered in prior training or validation stages.

\subsection{Evaluation Metrics}

We follow the standard protocol~\cite{soundspaces} and adopt three metrics to compare the navigation performance of different methods: (1) Success weighted by Path Length (SPL)~\ref{eq:spl}: After an agent successfully navigates to the target sound source, this metric measures how close the agent’s actual travel path is to the shortest feasible path~\cite{SPL}; (2) Success Rate (SR)~\ref{eq:sr}: The proportion of test samples in which the agent successfully reaches the target location; (3) Success-Navigation Accuracy (SNA)~\ref{eq:sna}: This metric evaluates the agent’s ability to not only successfully reach the target during navigation but also maintain orientation toward the correct direction.

\begin{equation}
  \text{SR} = \frac{1}{N} \sum_{i=1}^{N} S_i,
  \label{eq:sr}
\end{equation}

\begin{equation}
  \text{SNA} = \frac{1}{N} \sum_{i=1}^{N} S_i \cdot \frac{l_i}{a_i}~\cite{Waypoint},
  \label{eq:sna}
\end{equation}

\begin{equation}
  \text{SPL} = \frac{1}{N} \sum_{i=1}^{N} S_i \cdot \frac{l_i}{\max(p_i, l_i)}~\cite{SPL}.
  \label{eq:spl}
\end{equation}

In the formulas, \(N\) is the total number of test episodes; \(S_i \in \{0,1\}\) specifies whether the \(i\)-th episode is successful; \(l_i\) is the shortest feasible path length for episode \(i\); \(p_i\) is the actual path length traveled by the agent; and \(a_i\) is the number of actions executed in episode \(i\). The term \(a_i\) includes inefficient operations such as in-place rotations, so \(\text{SNA}\) penalizes excessive redundant movements.

\begin{figure*}
    \centering
    \includegraphics[width=1\linewidth]{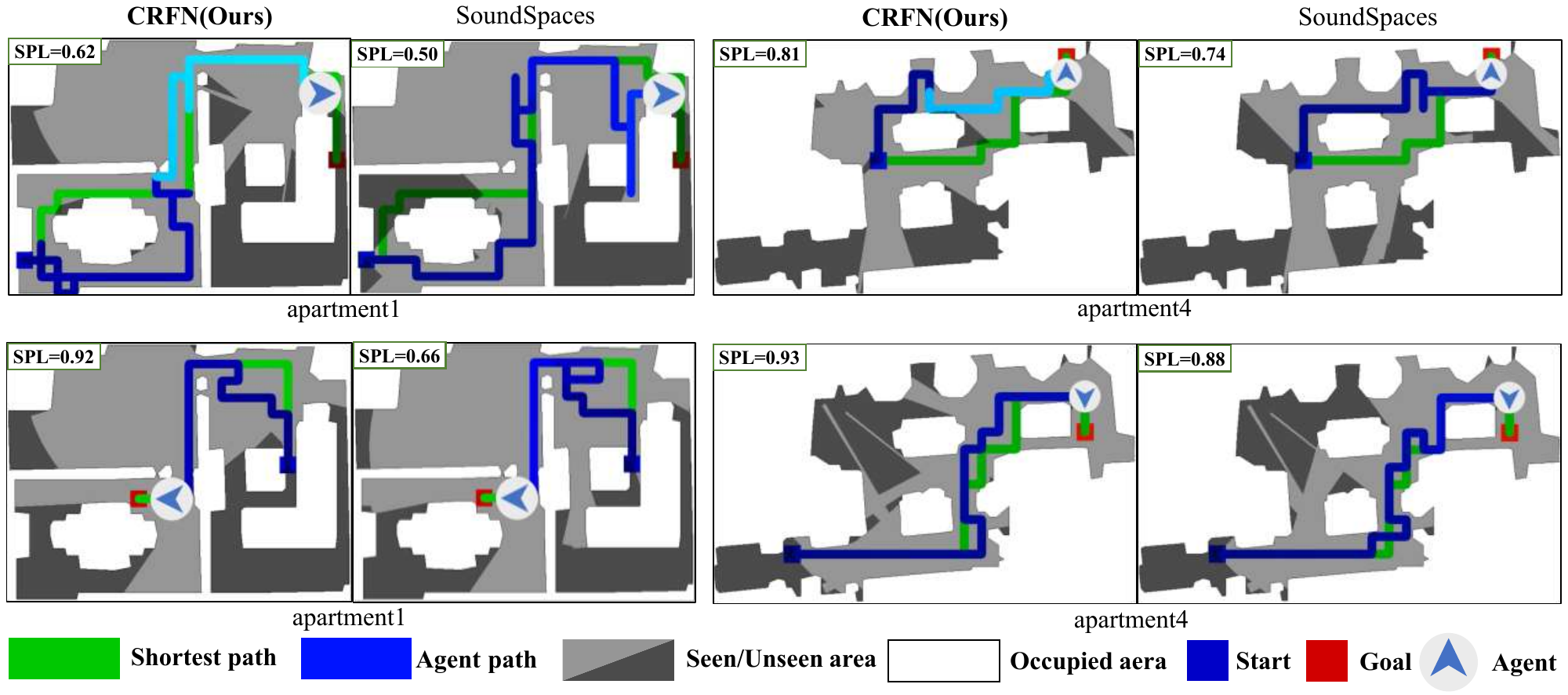}
    \caption{Navigation trajectories on the top-down map in the Replica scenes.~Agent paths transition from dark to light blue temporally, while green indicates the shortest geodesic path. }
    \label{guiji1}
\end{figure*}

\begin{figure*}
    \centering
    \includegraphics[width=1\linewidth]{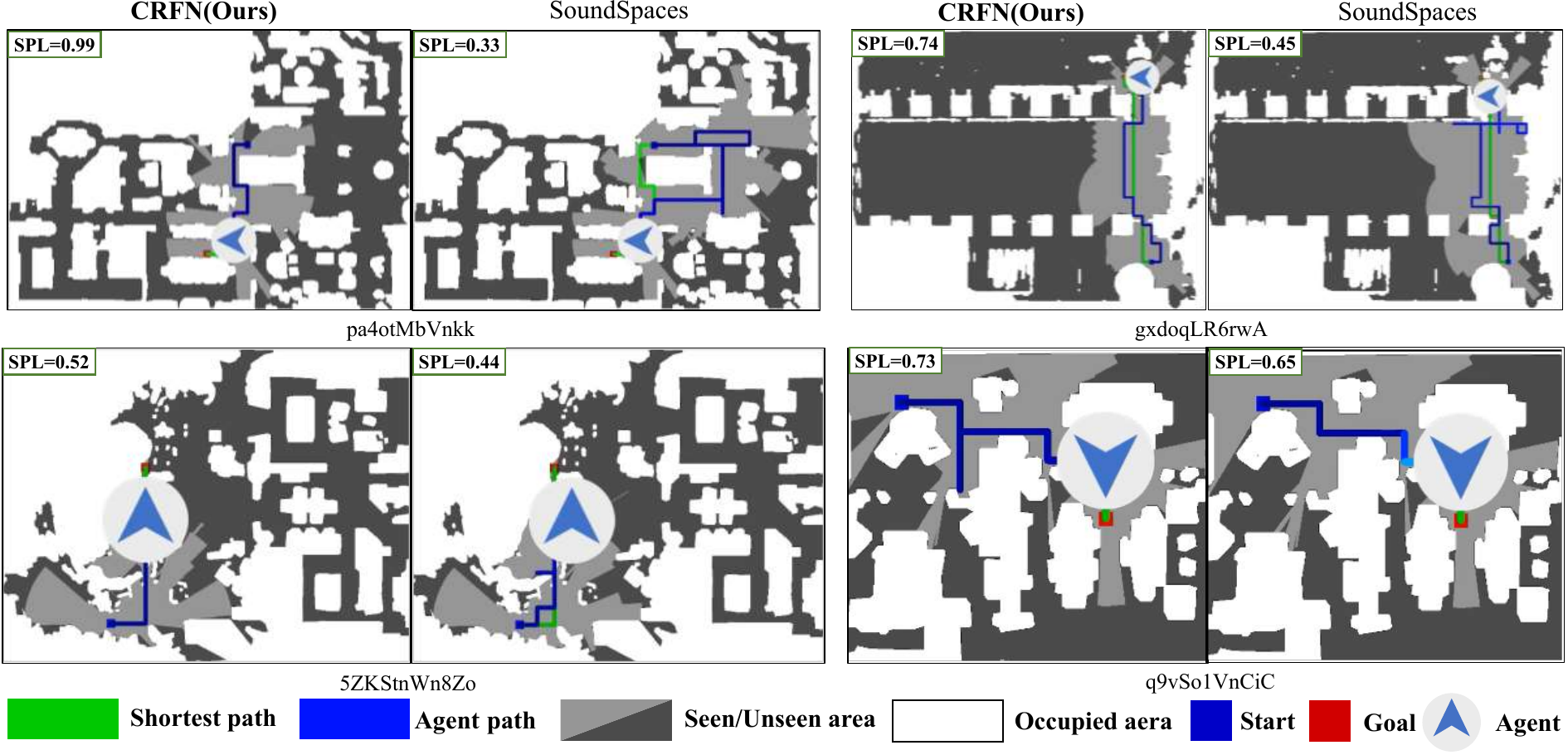}
    \caption{Navigation trajectories on the top-down map in the Matterport3D scenes.~Agent paths transition from dark to light blue temporally, while green indicates the shortest geodesic path.}
    \label{guiji2}
\end{figure*}

\subsection{Baselines}

We compare our method with the following approaches:
\begin{itemize}
    \item \textbf{Random Agent}: this is the simplest task in navigation, where the agent randomly selects actions (turn left, turn right, move forward) at each time step to search for the target.

    \item \textbf{Direction Follower}: this method adopts a hierarchical navigation framework, where the high-level policy predicts the Direction of Arrival of the sound source relying solely on auditory signals, and sets a waypoint at a fixed distance (K meters) along this direction. After the agent reaches the current waypoint, it repeats this process to gradually approach the target.

    \item \textbf{Frontier Waypoints}: the hierarchical baseline selects the next waypoint at the intersection of the predicted DoA and the current exploration frontier. Frontier-based waypoints are standard in visual navigation~\cite{baselineFW1,baselineFW2,baselineFW3}, making this a representative baseline of standard practice.

    \item \textbf{Supervised Waypoints}: the hierarchical baseline takes the RGB image and audio spectrogram as inputs and, using supervised (non-end-to-end) training, predicts waypoints within the field of view. Its design follows the supervised waypoint predictor of Bansal et al~\cite{bansal2020}.

    \item \textbf{Gan et al.}~\cite{gan2020look} : one AVN method developed based on the AI2-THOR platform~\cite{AI2-THOR}. As a representative work in the early stage of the AVN field, its core advantage lies in the first systematic implementation of collaborative modeling between visual spatial memory and auditory sound source localization, providing a benchmark paradigm for subsequent related research.

    \item \textbf{SoundSpaces}~\cite{soundspaces}: the first publicly available audio-visual embodied navigation simulation platform. Our work is based on their code.

\end{itemize}

\begin{figure*}[ht!]
    \centering
    \includegraphics[width=1\linewidth]{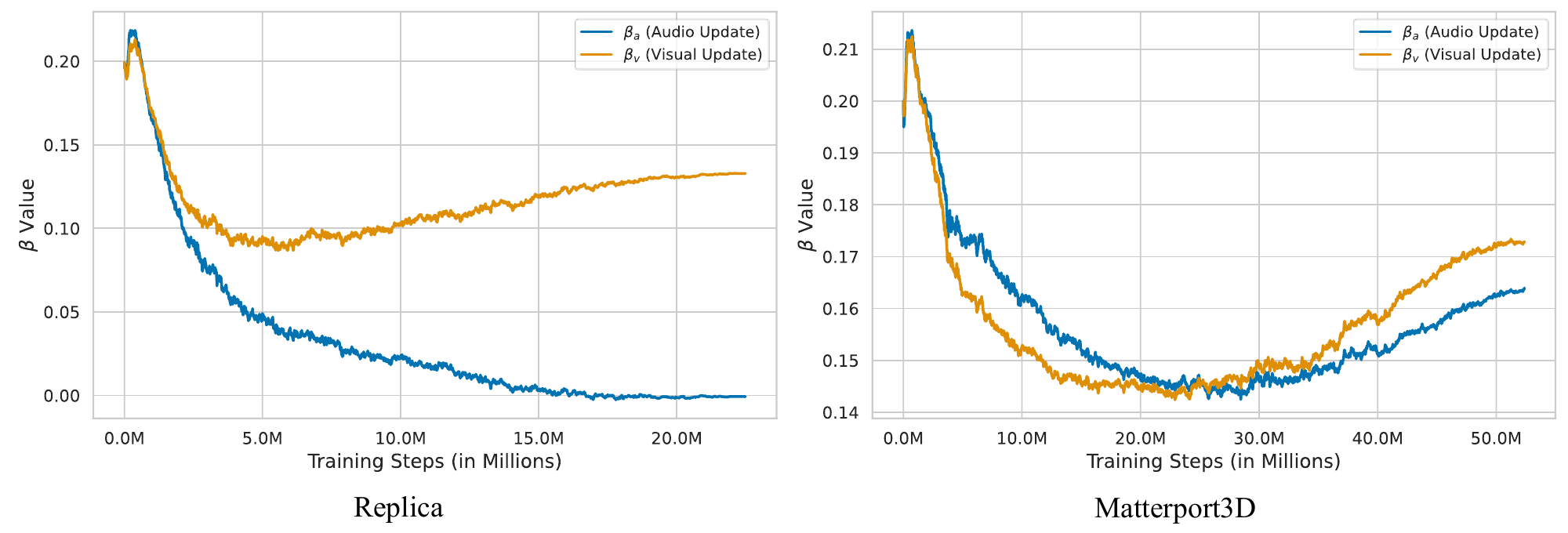}
    \caption{\textbf{Variation of Modal Weights with Training Progress.} In the figure, the curves represent the evolutionary trends of the visual update weight (\(\beta_v\)) and auditory update weight (\(\beta_a\)) on the Replica (left) and Matterport3D (right) datasets, respectively.}
    \label{beat}
\end{figure*}

\subsection{Quantitative Experimental Results}
Table~\ref{tab:depth} presents a comparison of the proposed method (CRFN) with representative baselines on the Replica~\cite{replica} and Matterport3D~\cite{matterport3d} datasets under the Depth setting. As shown, CRFN consistently achieves the best performance across all evaluation metrics, demonstrating its effectiveness in cross-modal feature integration and robust audio-visual navigation. On the Replica dataset, CRFN surpasses all prior methods in both Heard and Unheard settings. Specifically, in the Heard scenario, the SPL improves from 74.4 to 76.7, and SR rises to 93.1, indicating that the residual cross-modal fusion effectively enhances audio-visual synergy. In the Unheard scenario, the improvement becomes more pronounced---SPL reaches 41.6, about 7.6 percentage points higher than SoundSpaces~\cite{soundspaces}---showing stronger generalization and robustness to unseen sound sources. Likewise, on the more complex Matterport3D dataset, CRFN achieves the highest scores in both settings, with SPL = 57.3 and SR = 70.3 in the Heard case, significantly outperforming existing approaches. These results confirm that CRFN better balances the complementary relationship between visual and auditory signals in realistic environments, leading to improved multimodal perception, decision-making, and generalization in unseen scenes.

\subsection{Qualitative Experimental Results}

Fig.~\ref{guiji1} and~\ref{guiji2} provide a visual comparison of the top-down navigation trajectories between our CRFN model and the SoundSpaces baseline across a series of typical scenes. Specifically, Fig.~\ref{guiji1} showcases two representative scenes from the Replica dataset, while Fig.~\ref{guiji2} presents four complex scenes from the Matterport3D~\cite{matterport3d} dataset. From these trajectories, it can be clearly observed that the navigation paths generated by CRFN are significantly superior to those of the baseline, whether in the structurally simpler scenes of Replica (e.g., \textit{apartment1} and \textit{apartment4} in Fig.~\ref{guiji1}) or in the more complex layouts of Matterport3D~\cite{matterport3d}. CRFN's trajectories adhere more closely to the shortest path (green), exhibiting less invalid exploration and backtracking, which is also corroborated by its higher SPL scores. Particularly in the task with scene ID \textit{pa4otMbVnkk}, CRFN almost perfectly replicates the shortest path (SPL=0.99), whereas the SoundSpaces~\cite{soundspaces} baseline, unable to establish a stable audio-visual correspondence, wanders extensively and inefficiently within the environment, resulting in extremely low path efficiency (SPL=0.33). This indicates that our model can efficiently fuse cross-modal information to quickly lock onto the target's bearing even in acoustically simple, open spaces, thereby suppressing aimless exploration. Furthermore, in scenarios like \textit{apartment4}, where the sound source may be partially occluded by walls, CRFN is still able to progressively correct its course and successfully complete the navigation, while the baseline model exhibits greater uncertainty, leading to lower path efficiency.

\begin{table}
\centering
\caption{Ablation study on the Fusion Controller of CRFN on the Replica dataset.}
\label{xr1}
\renewcommand{\arraystretch}{1.1}
\setlength{\tabcolsep}{2pt}        
\small
\begin{tabular}{l|ccc|ccc}
\toprule
\multirow{2}{*}{\textbf{Method}}  
  & \multicolumn{3}{c|}{\textbf{Telephone}}      
  & \multicolumn{3}{c}{\textbf{Unheard}} \\
\cmidrule(lr){2-4} \cmidrule(lr){5-7}
& \textbf{SPL} $\uparrow$ & \textbf{SR} $\uparrow$ & \textbf{SNA} $\uparrow$
& \textbf{SPL} $\uparrow$ & \textbf{SR} $\uparrow$ & \textbf{SNA} $\uparrow$ \\
\midrule
w/o FC           & 69.3 & 84.9 & 44.4 & 34.0 & 43.5 & 19.2 \\
\textbf{CRFN (Ours)} & 76.7 & 93.1 & 47.3 & 41.6 & 55.7 & 22.5 \\
\bottomrule
\end{tabular}
\end{table}

\begin{table}
\centering
\caption{Ablation study on the Fusion Controller of CRFN on the Matterport3D dataset.}
\label{xr2}
\renewcommand{\arraystretch}{1.1}
\setlength{\tabcolsep}{2pt}      
\small
\begin{tabular}{l|ccc|ccc}
\toprule
\multirow{2}{*}{\textbf{Method}}  
  & \multicolumn{3}{c|}{\textbf{Telephone}}      
  & \multicolumn{3}{c}{\textbf{Unheard}} \\
\cmidrule(lr){2-4} \cmidrule(lr){5-7}
& \textbf{SPL} $\uparrow$ & \textbf{SR} $\uparrow$ & \textbf{SNA} $\uparrow$
& \textbf{SPL} $\uparrow$ & \textbf{SR} $\uparrow$ & \textbf{SNA} $\uparrow$ \\
\midrule
w/o FC           & 53.1 & 65.9 & 32.2 & 28.9 & 37.7 & 14.7 \\
\textbf{CRFN (Ours)} & 57.3 & 70.3 & 33.2 & 27.7 & 40.1 & 13.5 \\
\bottomrule
\end{tabular}
\end{table}

\subsection{Modal Dependence Analysis}

To further reveal the internal learning dynamics of our model across different environments, we visualize the temporal evolution of the update weights for the visual ($\beta_v$) and auditory ($\beta_a$) modalities during training. As shown in Fig.~\ref{beat}, we observe a notable phenomenon: rather than fusing multimodal information in a static manner, the model adaptively adjusts its modality dependency according to the structural complexity and perceptual statistics of the environment, exhibiting a dynamic balancing strategy between modalities.

Specifically, on the Replica~\cite{replica} dataset, the model progressively reduces its reliance on the auditory modality ($\beta_a$) in the later stages of training, eventually converging to a visually dominated navigation strategy. This behavior can be attributed to the relatively small scale and regular structure of Replica~\cite{replica} environments, where the agent can easily memorize spatial layouts through visual cues. Once the visual signal alone becomes sufficient for localization and planning, the model naturally suppresses the contribution of the temporally uncertain audio input, leading to a more efficient unimodal decision process.

In contrast, on the Matterport3D~\cite{matterport3d} dataset, the model maintains a balanced dependence on both modalities throughout training. Matterport3D~\cite{matterport3d} contains large-scale, cluttered real-world scenes with complex geometry, occlusions, and reverberations, where a single modality provides insufficient information for reliable navigation. In this case, the audio modality offers complementary global directional cues that can penetrate obstacles, compensating for the limitations of visual perception. This results in a more robust cross-modal collaboration, highlighting the model’s inherent ability to adaptively modulate modality weights across diverse environments.

\subsection{Ablation Studies}
To evaluate the effectiveness of the Fusion Controller (FC) in our CRFN, we conducted ablation experiments on both the Replica~\cite{replica} and Matterport3D~\cite{matterport3d} datasets. As shown in Tables~\ref{xr1} and~\ref{xr2}, removing the Fusion Controller (w/o FC) leads to a clear performance drop across all metrics. On the Replica~\cite{replica} dataset, the SPL decreases from 76.7 to 69.3 and SR from 93.1 to 84.9, indicating that the absence of adaptive fusion regulation causes imbalance between modalities and degrades navigation efficiency in structured synthetic scenes. On the Matterport3D~\cite{matterport3d} dataset, performance also declines, though to a lesser extent, suggesting that in complex real-world environments the model can still partially rely on natural audio-visual complementarity. Overall, these results demonstrate that the Fusion Controller plays a crucial role in stabilizing cross-modal interactions, mitigating modality dominance, and enhancing generalization across diverse environments.

In this experiment, we investigate the effect of the initial residual scaling factor \(\beta_{\text{init}}\) on the performance of the proposed model. The residual scaling factor controls the initial strength of the cross-modal residual connections in CRFN, which is crucial for balancing the contributions from both audio and visual modalities during training. We test three different values of \(\beta_{\text{init}}\), namely \(0.1\), \(0.2\), and \(0.3\), across two popular datasets, Replica and Matterport3D. The results, summarized in Table~\ref{xr3}, show that a moderate value of \(\beta_{\text{init}} = 0.2\) leads to the best overall performance, achieving higher SPL, SR, and SNA across both datasets. This suggests that initializing the residual scaling factor with a moderate value helps ensure a stable learning process, avoiding overdominance of any single modality while maintaining effective cross-modal fusion.

\section{Conlusion}

\begin{table}
\centering
\caption{Effect of the residual scaling initialization factor $\beta_{\text{init}}$ on navigation performance. Results show that a moderate initialization ($\beta_{\text{init}}=0.2$) yields the best balance between training stability and accuracy on both Replica and Matterport3D datasets.}
\label{xr3}
\renewcommand{\arraystretch}{1.1}
\setlength{\tabcolsep}{6pt}        
\small
\begin{tabular}{l|ccc|ccc}
\toprule
\multirow{2}{*}{\textbf{$\beta_{\text{init}}$}}  
  & \multicolumn{3}{c|}{\textbf{Replica}}      
  & \multicolumn{3}{c}{\textbf{Matterport3D}} \\
\cmidrule(lr){2-4} \cmidrule(lr){5-7}
& \textbf{SPL} $\uparrow$ & \textbf{SR} $\uparrow$ & \textbf{SNA} $\uparrow$
& \textbf{SPL} $\uparrow$ & \textbf{SR} $\uparrow$ & \textbf{SNA} $\uparrow$ \\
\midrule
0.1     & 76.4 & 91.6 & 46.4 & 53.2 & 64.4 & 31.3 \\
0.2     & 76.7 & 93.1 & 47.3 & 57.3 & 70.3 & 33.2 \\
0.3     & 75.2 & 92.6 & 46.1 & 53.3 & 65.5 & 32.0 \\
\bottomrule
\end{tabular}
\end{table}

This paper presents CRFN, a cross-modal residual fusion network designed for audio-visual navigation tasks. CRFN achieves fine-grained alignment and complementary modeling between modalities through a bidirectional residual interaction mechanism, while a fusion controller dynamically adjusts the contribution of each modality, effectively suppressing imbalance and feature degradation. Experimental results demonstrate that CRFN achieves superior navigation performance and stable cross-domain generalization in diverse and complex environments, validating the effectiveness and robustness of the proposed fusion mechanism.

Moreover, by analyzing the dynamic evolution of residual coefficients, we uncover an interesting adaptive modality dependence phenomenon: agents tend to rely more on visual cues in high-fidelity synthetic environments, whereas cross-modal complementarity becomes crucial in complex real-world scenes. This finding provides new insights into how embodied agents coordinate multimodal perception and decision-making.

\section*{Acknowledgements}

This research was financially supported by the National Natural Science Foundation of China (Grant No. 62463029).


{\small
\bibliographystyle{cvm}
\bibliography{references}

@inproceedings{soundspaces,
  title={SoundSpaces: Audio-Visual Navigation in 3D Environments},
  author={Chen, Changan and Jain, Unnat and Schissler, Carl and Gari, Sebastià Vicenc Amengual and Al-Halah, Ziad and Ithapu, Vamsi Krishna and Robinson, Philip and Grauman, Kristen},
  booktitle={Proceedings of the IEEE/CVF Conference on Computer Vision and Pattern Recognition (CVPR)},
  pages={17--36},
  year={2020},
  publisher={IEEE/CVF},
}

@inproceedings{gan2020look,
  title={Look, listen, and act: Towards audio-visual embodied navigation},
  author={Gan, Chuang and Zhang, Yiwei and Wu, Jiajun and Gong, Boqing and Tenenbaum, Joshua B},
  booktitle={2020 IEEE International Conference on Robotics and Automation (ICRA)},
  pages={9701--9707},
  year={2020},
  organization={IEEE}
}

@inproceedings{matterport3d,
  title={Matterport3D: Learning from RGB-D data in indoor environments},
  author={Chang, Angel and Dai, Angela and Funkhouser, Thomas and Halber, Maciej and Niebner, Matthias and Savva, Manolis and Song, Shuran and Zeng, Andy and Zhang, Yinda},
  booktitle={7th IEEE International Conference on 3D Vision, 3DV 2017},
  pages={667--676},
  year={2018},
  organization={Institute of Electrical and Electronics Engineers Inc.}
}

@article{replica,
  title={The replica dataset: A digital replica of indoor spaces},
  author={Straub, Julian and Whelan, Thomas and Ma, Lingni and Chen, Yufan and Wijmans, Erik and Green, Simon and Engel, Jakob J and Mur-Artal, Raul and Ren, Carl and Verma, Shobhit and others},
  journal={arXiv preprint arXiv:1906.05797},
  year={2019}
}

@inproceedings{habitat,
  title={Habitat: A platform for embodied ai research},
  author={Savva, Manolis and Kadian, Abhishek and Maksymets, Oleksandr and Zhao, Yili and Wijmans, Erik and Jain, Bhavana and Straub, Julian and Liu, Jia and Koltun, Vladlen and Malik, Jitendra and others},
  booktitle={Proceedings of the IEEE/CVF international conference on computer vision},
  pages={9339--9347},
  year={2019}
}

@article{habitat2.0,
  title={Habitat 2.0: Training home assistants to rearrange their habitat},
  author={Szot, Andrew and Clegg, Alexander and Undersander, Eric and Wijmans, Erik and Zhao, Yili and Turner, John and Maestre, Noah and Mukadam, Mustafa and Chaplot, Devendra Singh and Maksymets, Oleksandr and others},
  journal={Advances in neural information processing systems},
  volume={34},
  pages={251--266},
  year={2021}
}

@inproceedings{SAVi,
  title={Semantic audio-visual navigation},
  author={Chen, Changan and Al-Halah, Ziad and Grauman, Kristen},
  booktitle={Proceedings of the IEEE/CVF Conference on Computer Vision and Pattern Recognition},
  pages={15516--15525},
  year={2021}
}

@inproceedings{Waypoint,
  title     = {Learning to Set Waypoints for Audio-Visual Navigation},
  author    = {Chen, Changan and Majumder, Sagnik and Al-Halah, Ziad and Gao, Ruohan and Ramakrishnan, Santhosh Kumar and Grauman, Kristen},
  booktitle = {9th International Conference on Learning Representations, ICLR 2021},
  year      = {2021},
  address   = {Virtual Event, Austria},
  month     = {May 3--7},
  publisher = {OpenReview.net},
}

@article{catchme,
  title={Catch me if you hear me: Audio-visual navigation in complex unmapped environments with moving sounds},
  author={Younes, Abdelrahman and Honerkamp, Daniel and Welschehold, Tim and Valada, Abhinav},
  journal={IEEE Robotics and Automation Letters},
  volume={8},
  number={2},
  pages={928--935},
  year={2023},
  publisher={IEEE}
}

@inproceedings{SAAVN,
  author    = {Yinfeng Yu and
               Wenbing Huang and
               Fuchun Sun and
               Changan Chen and
               Yikai Wang and
               Xiaohong Liu},
  title     = {Sound Adversarial Audio-Visual Navigation},
  booktitle = {The Tenth International Conference on Learning Representations, {ICLR} 2022, Virtual Event, April 25-29, 2022},
  publisher = {OpenReview.net},
  year      = {2022}
}

@article{SPL,
  title={On evaluation of embodied navigation agents},
  author={Anderson, Peter and Chang, Angel and Chaplot, Devendra Singh and Dosovitskiy, Alexey and Gupta, Saurabh and Koltun, Vladlen and Kosecka, Jana and Malik, Jitendra and Mottaghi, Roozbeh and Savva, Manolis and others},
  journal={arXiv preprint arXiv:1807.06757},
  year={2018}
}

@inproceedings{FSAVVN,
  title     = {Pay Self-Attention to Audio-Visual Navigation},
  author    = {Yu, Yinfeng and Cao, Lele and Sun, Fuchun and Liu, Xiaohong and Wang, Liejun},
  booktitle = {33rd British Machine Vision Conference 2022, BMVC 2022},
  year      = {2022},
  address   = {London, UK},
  month     = {November},
  pages     = {46},
  publisher = {BMVA Press},
}

@inproceedings{bansal2020,
  title={Combining optimal control and learning for visual navigation in novel environments},
  author={Bansal, Somil and Tolani, Varun and Gupta, Saurabh and Malik, Jitendra and Tomlin, Claire},
  booktitle={Conference on Robot Learning},
  pages={420--429},
  year={2020},
  organization={PMLR}
}

@inproceedings{tjavn,
  title={Towards Audio-Visual Navigation in Noisy Environments: A Large-Scale Benchmark Dataset and an Architecture Considering Multiple Sound-Sources},
  author={Shi, Zhanbo and Zhang, Lin and Li, Linfei and Shen, Ying},
  booktitle={Proceedings of the AAAI Conference on Artificial Intelligence},
  volume={39},
  pages={14673--14680},
  year={2025}
}

@article{yuttt,
    author = {Yu, Yinfeng and Cao, Lele and Sun, Fuchun and Yang, Chao and Lai, Huicheng and Huang, Wenbing},
    title = {Echo-Enhanced Embodied Visual Navigation},
    journal = {Neural Computation},
    volume = {35},
    number = {5},
    pages = {958-976},
    year = {2023},
    month = {04},
    issn = {0899-7667}			   
 }

@inproceedings{yu2023measuring,
  title={Measuring Acoustics with Collaborative Multiple Agents},
  author={Yu, Yinfeng and Chen, Changan and Cao, Lele and Yang, Fangkai and Sun, Fuchun},
  booktitle = {Proceedings of the 32nd International Joint Conference on Artificial Intelligence (IJCAI-23)},
  year={2023},
  address={Macao, China},
  month={August}
}

@inproceedings{ORAN,
  title={Omnidirectional information gathering for knowledge transfer-based audio-visual navigation},
  author={Chen, Jinyu and Wang, Wenguan and Liu, Si and Li, Hongsheng and Yang, Yi},
  booktitle={Proceedings of the IEEE/CVF International Conference on Computer Vision},
  pages={10993--11003},
  year={2023}
}

@inproceedings{PointGN,
  title={Vision-and-language navigation: Interpreting visually-grounded navigation instructions in real environments},
  author={Anderson, Peter and Wu, Qi and Teney, Damien and Bruce, Jake and Johnson, Mark and S{\"u}nderhauf, Niko and Reid, Ian and Gould, Stephen and Van Den Hengel, Anton},
  booktitle={Proceedings of the IEEE conference on computer vision and pattern recognition},
  pages={3674--3683},
  year={2018}
}

@article{ObjectGN,
  title={Objectnav revisited: On evaluation of embodied agents navigating to objects},
  author={Batra, Dhruv and Gokaslan, Aaron and Kembhavi, Aniruddha and Maksymets, Oleksandr and Mottaghi, Roozbeh and Savva, Manolis and Toshev, Alexander and Wijmans, Erik},
  journal={arXiv preprint arXiv:2006.13171},
  year={2020}
}

@inproceedings{CAVEN,
  title={Caven: An embodied conversational agent for efficient audio-visual navigation in noisy environments},
  author={Liu, Xiulong and Paul, Sudipta and Chatterjee, Moitreya and Cherian, Anoop},
  booktitle={Proceedings of the AAAI conference on artificial intelligence},
  volume={38},
  pages={3765--3773},
  year={2024}
}

@inproceedings{RILA,
  title={Rila: Reflective and imaginative language agent for zero-shot semantic audio-visual navigation},
  author={Yang, Zeyuan and Liu, Jiageng and Chen, Peihao and Cherian, Anoop and Marks, Tim K and Le Roux, Jonathan and Gan, Chuang},
  booktitle={Proceedings of the IEEE/CVF Conference on Computer Vision and Pattern Recognition},
  pages={16251--16261},
  year={2024}
}

@inproceedings{visualN1,
  title={Cognitive mapping and planning for visual navigation},
  author={Gupta, Saurabh and Davidson, James and Levine, Sergey and Sukthankar, Rahul and Malik, Jitendra},
  booktitle={Proceedings of the IEEE conference on computer vision and pattern recognition},
  pages={2616--2625},
  year={2017}
}

@article{visualN2,
  title={Dd-ppo: Learning near-perfect pointgoal navigators from 2.5 billion frames},
  author={Wijmans, Erik and Kadian, Abhishek and Morcos, Ari and Lee, Stefan and Essa, Irfan and Parikh, Devi and Savva, Manolis and Batra, Dhruv},
  journal={arXiv preprint arXiv:1911.00357},
  year={2019}
}

@article{baselineFW1,
  title={Deep learning of structured environments for robot search},
  author={Caley, Jeffrey A and Lawrance, Nicholas RJ and Hollinger, Geoffrey A},
  journal={Autonomous Robots},
  volume={43},
  number={7},
  pages={1695--1714},
  year={2019},
  publisher={Springer}
}

@inproceedings{baselineFW2,
  title={Learning over subgoals for efficient navigation of structured, unknown environments},
  author={Stein, Gregory J and Bradley, Christopher and Roy, Nicholas},
  booktitle={Conference on robot learning},
  pages={213--222},
  year={2018},
  organization={PMLR}
}

@inproceedings{baselineFW3,
  title={Neural topological slam for visual navigation},
  author={Chaplot, Devendra Singh and Salakhutdinov, Ruslan and Gupta, Abhinav and Gupta, Saurabh},
  booktitle={Proceedings of the IEEE/CVF conference on computer vision and pattern recognition},
  pages={12875--12884},
  year={2020}
}

@article{AI2-THOR,
  title={Ai2-thor: An interactive 3d environment for visual ai},
  author={Kolve, Eric and Mottaghi, Roozbeh and Han, Winson and VanderBilt, Eli and Weihs, Luca and Herrasti, Alvaro and Deitke, Matt and Ehsani, Kiana and Gordon, Daniel and Zhu, Yuke and others},
  journal={arXiv preprint arXiv:1712.05474},
  year={2017}
}

@article{srivastava2012multimodal,
  title={Multimodal learning with deep boltzmann machines},
  author={Srivastava, Nitish and Salakhutdinov, Russ R},
  journal={Advances in neural information processing systems},
  volume={25},
  year={2012}
}

@article{kim2016multimodal,
  title={Multimodal residual learning for visual qa},
  author={Kim, Jin-Hwa and Lee, Sang-Woo and Kwak, Donghyun and Heo, Min-Oh and Kim, Jeonghee and Ha, Jung-Woo and Zhang, Byoung-Tak},
  journal={Advances in neural information processing systems},
  volume={29},
  year={2016}
}

@inproceedings{embodied1,
  title={Affordances-oriented planning using foundation models for continuous vision-language navigation},
  author={Chen, Jiaqi and Lin, Bingqian and Liu, Xinmin and Ma, Lin and Liang, Xiaodan and Wong, Kwan-Yee K},
  booktitle={Proceedings of the AAAI Conference on Artificial Intelligence},
  volume={39},
  pages={23568--23576},
  year={2025}
}

@inproceedings{embodied2,
  title={Webvln: Vision-and-language navigation on websites},
  author={Chen, Qi and Pitawela, Dileepa and Zhao, Chongyang and Zhou, Gengze and Chen, Hsiang-Ting and Wu, Qi},
  booktitle={Proceedings of the AAAI Conference on Artificial Intelligence},
  volume={38},
  pages={1165--1173},
  year={2024}
}

@article{embodied3,
  title={Embodied navigation with multi-modal information: A survey from tasks to methodology},
  author={Wu, Yuchen and Zhang, Pengcheng and Gu, Meiying and Zheng, Jin and Bai, Xiao},
  journal={Information Fusion},
  volume={112},
  pages={102532},
  year={2024},
  publisher={Elsevier}
}

@inproceedings{embodied4,
  title={Towards learning a generalist model for embodied navigation},
  author={Zheng, Duo and Huang, Shijia and Zhao, Lin and Zhong, Yiwu and Wang, Liwei},
  booktitle={Proceedings of the IEEE/CVF Conference on Computer Vision and Pattern Recognition},
  pages={13624--13634},
  year={2024}
}

@inproceedings{yu2025dope,
  title={DOPE: Dual Object Perception-Enhancement Network for Vision-and-Language Navigation},
  author={Yu, Yinfeng and Yang, Dongsheng},
  booktitle={Proceedings of the 2025 International Conference on Multimedia Retrieval},
  pages={1739--1748},
  year={2025}
}

@inproceedings{yu2021weavenet,
  title={Weavenet: End-to-end audiovisual sentiment analysis},
  author={Yu, Yinfeng and Jia, Zhenhong and Shi, Fei and Zhu, Meiling and Wang, Wenjun and Li, Xiuhong},
  booktitle={International Conference on Cognitive Systems and Signal Processing},
  pages={3--16},
  year={2021}
}
}

\end{document}